\documentclass[a4paper]{llncs}
\usepackage{amssymb}
\usepackage{graphicx}
\usepackage{amsmath}
\usepackage{url}
\usepackage{color}
\usepackage{multirow}
\usepackage{array}
\usepackage{booktabs}
\usepackage[colorlinks]{hyperref}
\usepackage[usenames,table]{xcolor}

\begin{document}

\mainmatter  

\title{CompNet: Complementary Segmentation Network for Brain MRI Extraction}


%
%
\author{Raunak Dey, Yi Hong}
%

\institute{Computer Science Department, University of Georgia}

%
%

\maketitle

\begin{abstract}

Brain extraction is a fundamental step for most brain imaging studies. In this paper, we investigate the problem of skull stripping and propose complementary segmentation networks (CompNets) to accurately extract the brain from T1-weighted MRI scans, for both normal and pathological brain images. The proposed networks are designed in the framework of encoder-decoder networks and have two pathways to learn features from both the brain tissue and its complementary part located outside of the brain. The complementary pathway extracts the features in the non-brain region and leads to a robust solution to brain extraction from MRIs with pathologies, which do not exist in our training dataset. We demonstrate the effectiveness of our networks by evaluating them on the OASIS dataset, resulting in the state of the art performance under the two-fold cross-validation setting. Moreover, the robustness of our networks is verified by testing on images with introduced pathologies and by showing its invariance to unseen brain pathologies. In addition, our complementary network design is general and can be extended to address other image segmentation problems with better generalization.

\end{abstract}

\section{Introduction}

Image segmentation aims to locate and extract objects of interest from an image, which is one of the fundamental problems in medical research. Take the brain extraction problem as an example. To study the brain, magnetic resonance imaging (MRI) is the most popular modality choice. However, before the quantitative analysis of brain MRIs, e.g., measuring normal brain development and degeneration, uncovering brain disorders such as Alzheimer's disease, or diagnosing brain tumors or lesions, skull stripping is typically a preliminary but essential step, and many approaches have been proposed to tackle this problem.

In literature, the approaches developed for brain MRI extraction can be divided into two categories: traditional methods (manual, intensity or shape model based, hybrid, and PCA-based methods \cite{han2017brain,souza2017open}) and deep learning methods~\cite{kleesiek2016deep,salehi2017auto}. Deep neural networks have demonstrated the improved quality of the predicted brain mask, compared to traditional methods. However, these deep networks focus on learning image features mainly for brain tissue from a training dataset, which is typically a collection of normal (or apparently normal) brain MRIs, because these images are more commonly available than brain scans with pathologies. Thus, their model performance is sensitive to unseen pathological tissues. 

In this paper, we propose a novel deep neural network architecture for skull stripping from brain MRIs, which improves the performance of existing methods on brain extraction and more importantly is invariant to brain pathologies by only training on publicly available regular brain scans. In our new design, a network learns features for both brain tissue and non-brain structures, that is, we consider the complementary information of an object that is outside of the region of interest in an image. For instance, the structures outside of the brain, e.g., the skull, are highly similar and consistent among the normal and pathological brain images. Leveraging such complementary information about the brain can help increase the robustness of a brain extraction method and enable it to handle images with unseen structures in the brain. 

\begin{figure}[t]
\includegraphics[width=1.0\columnwidth]{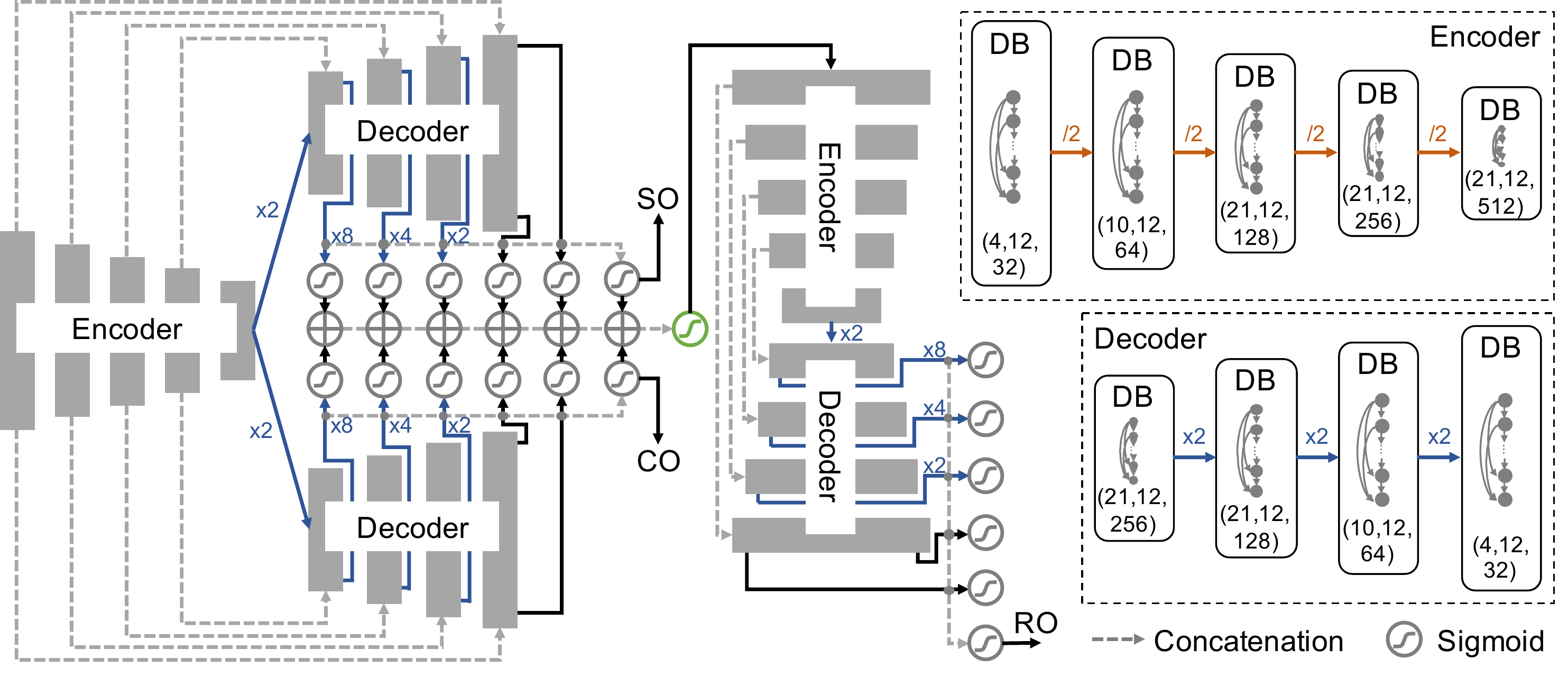}
\caption{Architecture of our complementary segmentation network, the optimal CompNet. The dense blocks (DB), corresponding to the gray bars, are used in each encoder and decoder. The triple ($x$,$y$,$z$) in each dense block indicates that it has $x$ convolutional layers with a kernel size $3\times3$; each layer has $y$ filters, except for the last one that has $z$ filters. SO: segmentation output for the brain mask; CO: complementary segmentation output for the non-brain mask; RO: reconstruction output for the input image. These three outputs produced by the Sigmoid function are the final predictions; while all other Sigmoids produce intermediate outputs, except for the green one that generates input for the image reconstruction sub-network. Best viewed in color.}
\label{fig:compnet}
\end{figure}

We explore multiple complementary segmentation networks (CompNets). In general, these networks have two pathways in common: one to learn what is the brain tissue and to generate a brain mask; the other to learn what is outside of the brain and to help the other branch generate a better brain mask. There are three variants, i.e., the probability, plain, and optimal CompNets. In particular, the probability CompNet needs an extra step to generate the ground truth for the complementary part such as the skull, while the plain and optimal CompNets do not need this additional input. The optimal CompNet is built upon the plain one and introduces dense blocks (a series of convolutional layers fully connected to each other~\cite{huang2017densely}) and multiple intermediate outputs~\cite{dey2018diagnostic}, as shown in Fig.~\ref{fig:compnet}. This optimal CompNet has an end-to-end design, fewer number of parameters to estimate, and the best performance among the three CompNets on both normal and pathological images from the OASIS dataset. In addition, this network is generic and can be applied in image segmentation if the complementary part of an object contributes to the understanding of the object in the image.

\section{CompNets: Complementary Segmentation Networks}
\label{sec:compnet}

An encoder-decoder network, like U-Net~\cite{ronneberger2015u}, is often used in image segmentation. Current segmentation networks mainly focus on objects of interest, which may lead to the difficulty in its generalization to unseen image data. In this section, we introduce our novel complementary segmentation networks (short for CompNet), which increase the segmentation robustness by incorporating the learning process of the object of interest with the learning of its complementary part in the image.


The architecture of our optimal CompNet is depicted in Fig.~\ref{fig:compnet}. This network has three components. The first component is a segmentation branch for the region of interest (ROI) such as the brain, which follows the U-Net architecture and generates the brain mask. Given normal brain scans, a network with only this branch focuses on extracting features of standard brain tissue, resulting in its difficulty of handling brain scans with unseen pathologies. To tackle this problem, we augment the segmentation branch by adding a complementary one to learn structures in the non-brain region, because they are relatively consistent among normal and pathological images. However, due to the lack of true masks for the complementary part, we adopt a sub-encoder-decoder network to reconstruct the input brain scan based on the outputs of the provious two branches. This third component guides the learning process of the complementary branch, similar to the unsupervised W-Net~\cite{xia2017w}.  It provides direct feedback to the segmentation and complementary branches and expects reasonable predictions from them as input to reconstruct the original input image. The complementary branch indirectly affects the segmentation branch since they share the encoder to extract features.


The optimal CompNet includes dense blocks and multiple intermediate outputs, which help reduce the number of parameters to estimate and make the network easier to optimize. For readability and a better understanding, we start with a discussion of the plain version in detail.

\paragraph{\bf Plain CompNet.} The plain network is a simplified version of the network shown in Fig.~\ref{fig:compnet}. Similar to U-Net, the encoder and decoder blocks (the gray bars in Fig.~\ref{fig:compnet}) of the segmentation and reconstruction sub-networks have two convolutional layers in each, with a kernel of size $3\times3$. The number of convolutional filters in the encoder starts from 32, followed by 64, 128, 256, and 512, while the number in the decoder starting from 256, followed by 128, 64, and 32. Each convolutional layer is followed by batch normalization~\cite{ioffe2015batch} and dropout~\cite{srivastava2014dropout}. After each gray bar in the encoder, the feature maps are downsampled by 2 using max pooling; while for the decoder the feature maps are upsampled by 2 using deconvolutional layers. Each segmentation branch of this plain network has only one final output from the last layer of the decoder, after applying the Sigmoid function. The two outputs of the segmentation branches are combined through addition and passed as input to the reconstruction sub-network. In this sub-network, the output from the last layer of the decoder is the reconstructed image. Like U-Net, we have the concatenation of feature maps from an encoder to its decoder at the same resolution level, shown by the gray arrows in Fig.~\ref{fig:compnet}.  

We use the Dice coefficient ($Dice(A, B) = 2|A\cap B|/(|A|+|B|)$~\cite{dice1945measures}) in the objective function to measure the goodness of segmentation predictions and the mean squared error (MSE) to measure the goodness of the reconstruction. In particular, the learning goal of this network is to maximize the Dice coefficient between the predicted mask for the ROI ($\hat{Y}_S$) and its ground truth ($Y_S$), minimize the Dice coefficient between the predicted mask for the non-ROI ($\hat{Y}_C$) and the ROI true mask, and minimize the MSE between the reconstructed image ($\hat{X}_R$) and the input image ($X$). We formulate the loss function for one sample as
\begin{equation}
Loss(Y_S, \hat{Y}_S, \hat{Y}_C, X, \hat{X}_R) = -Dice(Y_S, \hat{Y}_S) + Dice(Y_S, \hat{Y}_C) + MSE(X, \hat{X}_R). \label{eq:loss}
\end{equation}
Here, the reconstruction loss ensures that the complementary output is not an empty image and that the summation of segmentation and complementary outputs is not an entirely white image, because such inputs without a whole brain and skull structure map will result in a substantial reconstruction error. 

\paragraph{\bf Optimal CompNet.} The plain CompNet has nearly 18 million parameters. Introducing dense connections among convolutional layers can considerably reduce the number of parameters of a network and mitigate the vanishing gradient problem in a deep neural network. Therefore, we replace each gray block in the plain CompNet with a dense block, as shown in Fig.~\ref{fig:compnet}. Each dense block has different numbers of convolutional layers and filters. Specifically, the dense blocks in each encoder have 4, 10, 21, 21, and 21 convolutional layers, respectively, and the ones in each decoder have 21, 21, 10, and 4 layers, respectively. All these convolutional layers use the same kernel size $3\times3$ and the number of convolutional filters is $12$ in each layer, except for the last one, changing from 32, to 64, to 128, to 256, and to 512 in the five dense blocks of the encoder while changing from 256, to 128, to 64, and to 32 in the four dense blocks of the decoder. This design aims to increase the amount of information that is transferred from one dense block to its next one by using more feature maps. In addition, we place dropout at the transition between dense blocks. Through adopting these dense blocks, our optimal CompNet becomes much deeper while having fewer parameters (15.3 million) to optimize, compared to the plain one.


Another change made to the plain CompNet is introducing multiple intermediate outputs~\cite{dey2018diagnostic}. These early outputs can mitigate the vanishing gradient problem in a deep neural network by shorting the distance from the input to the output. As shown in Fig.~\ref{fig:compnet}, each decoder in the segmentation and reconstruction sub-networks has six outputs, one after each Sigmoid function. The first five outputs are intermediate outputs, which are generated from the original and upsampled feature maps of the first convolutional layer in each dense block of the decoder and the feature maps of the last convolutional layer in the last dense block. We observe that having an intermediate output at the beginning of each dense-block provides better performance than having it at the end. An extra one at the end of the last dense-block allows collecting features learned by this block. The concatenation of all feature maps used for the intermediate outputs generates the sixth output, which is the final output of that branch for prediction. Furthermore, we use addition operations to integrate each pair of intermediate or final outputs from the two segmentation branches and then use the concatenation operation to collect all of them, resulting in the input for the reconstruction sub-network via a Sigmoid function (the green one in Fig.~\ref{fig:compnet}). Each Sigmoid layer produces a feature map with one channel using a 1x1x1 convolutional filter, which normalizes its response value within $[0,1]$.


\paragraph{\bf Probability CompNet.} The reconstruction sub-network is to guide the learning process of the complementary pathway. One might replace it by providing the ground truth of the complementary part for training, e.g., generating the skull mask. This strategy is our first attempt, and it could be non-trivial for images with a noisy background. After having the true masks for both brain and non-brain regions, we can build a network containing only the segmentation and complementary branches in Fig.~\ref{fig:compnet}, by removing the reconstruction component. To leverage the complementary information, we build connections between the convolutional layers of the two branches at the same resolution level. In particular, the feature maps of a block from one segmentation branch are converted to a probability map, which is inverted and multiplied to the feature maps at the same resolution level of the other branch. We perform the same operations on the other branch. Essentially, one branch informs the other to focus on learning features of its complementary part. This network can also handle brain extraction from pathological images; however, both brain and skull masks are needed for training, and the image background noise will influence the result. Although we can set an intensity threshold to denoise the background, this hyper-parameter may vary among images collected from different brain MRI scanners.

\section{Experiments} 
\label{sec:results}

\paragraph{\bf Datasets.} 
We evaluate CompNets on the OASIS dataset~\cite{marcus2007open}, which consists of a collection of T1-weighted brain MRI scans of 416 subjects aged 18 to 96 and 100 of them clinically diagnosed mild to moderate Alzheimer's disease. We use a subset with 406 subjects that have both brain images and masks available, with image dimension of $256\times256\times256$. These subjects are randomly shuffled and equally divided into two chunks for training and testing with two-fold cross-validation, similar to~\cite{kleesiek2016deep} for comparison on (apparently) normal brain images. 
 
\begin{figure}[t]
\centering
\begin{tabular}{ccccccc}
& \begin{tabular}{c} \small{Input} \\ \small{Image} \end{tabular} &
\begin{tabular}{c} \small{Plain} \\ \small{U-Net} \end{tabular} &
\begin{tabular}{c} \small{Dense} \\ \small{U-Net} \end{tabular} &
\begin{tabular}{c} \small{Prob.} \\ \small{CompNet} \end{tabular} &
\begin{tabular}{c} \small{Plain} \\ \small{CompNet} \end{tabular} &
\begin{tabular}{c} \small{Optimal} \\ \small{CompNet} \end{tabular} \\
\small{Normal} &
\raisebox{-.5\height}{\includegraphics[width=0.142\columnwidth]{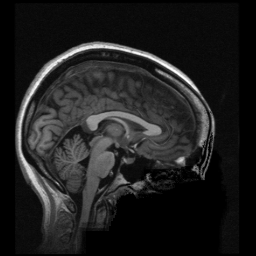}} &
\raisebox{-.5\height}{\includegraphics[width=0.142\columnwidth]{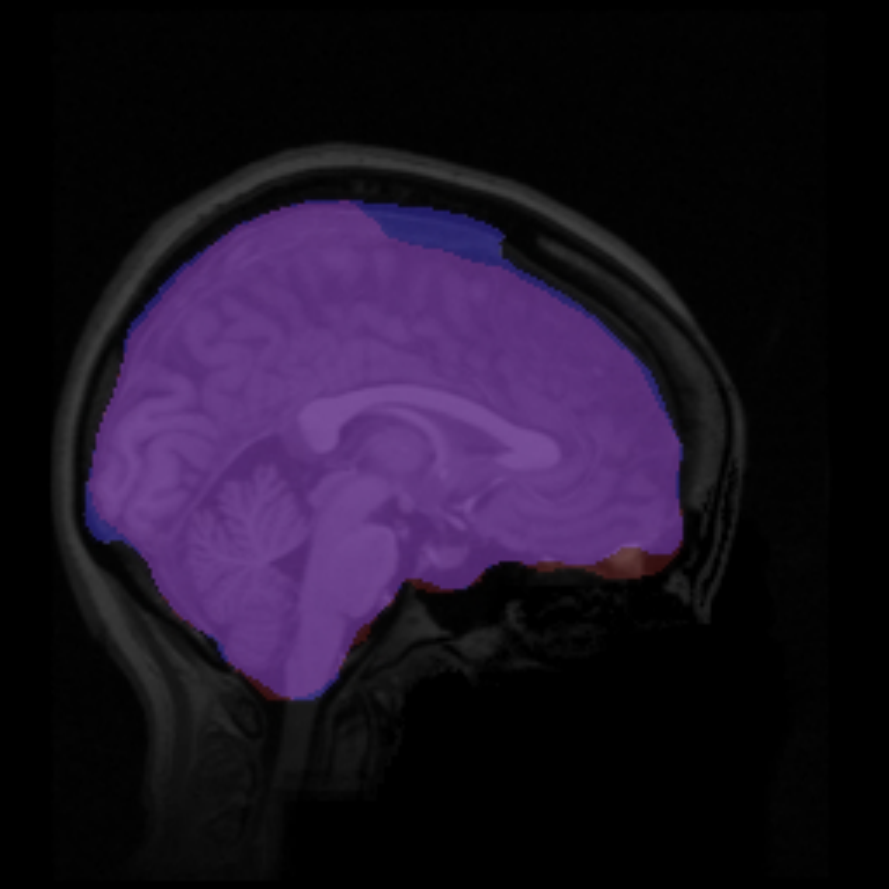}} &
\raisebox{-.5\height}{\includegraphics[width=0.142\columnwidth]{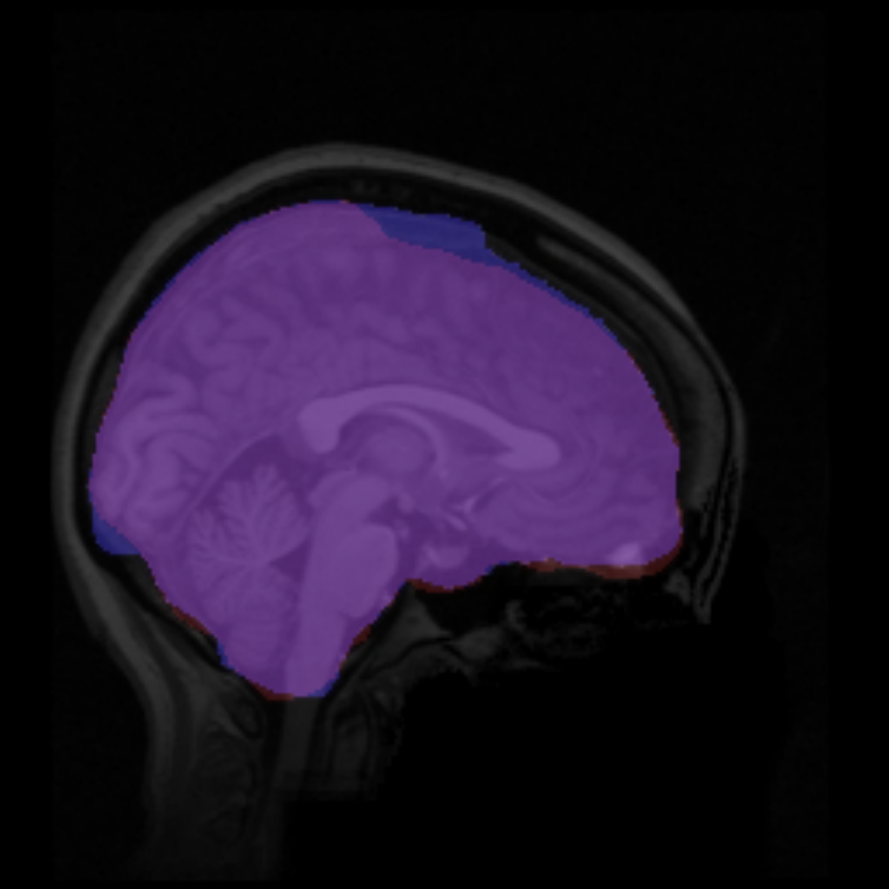}} &
\raisebox{-.5\height}{\includegraphics[width=0.142\columnwidth]{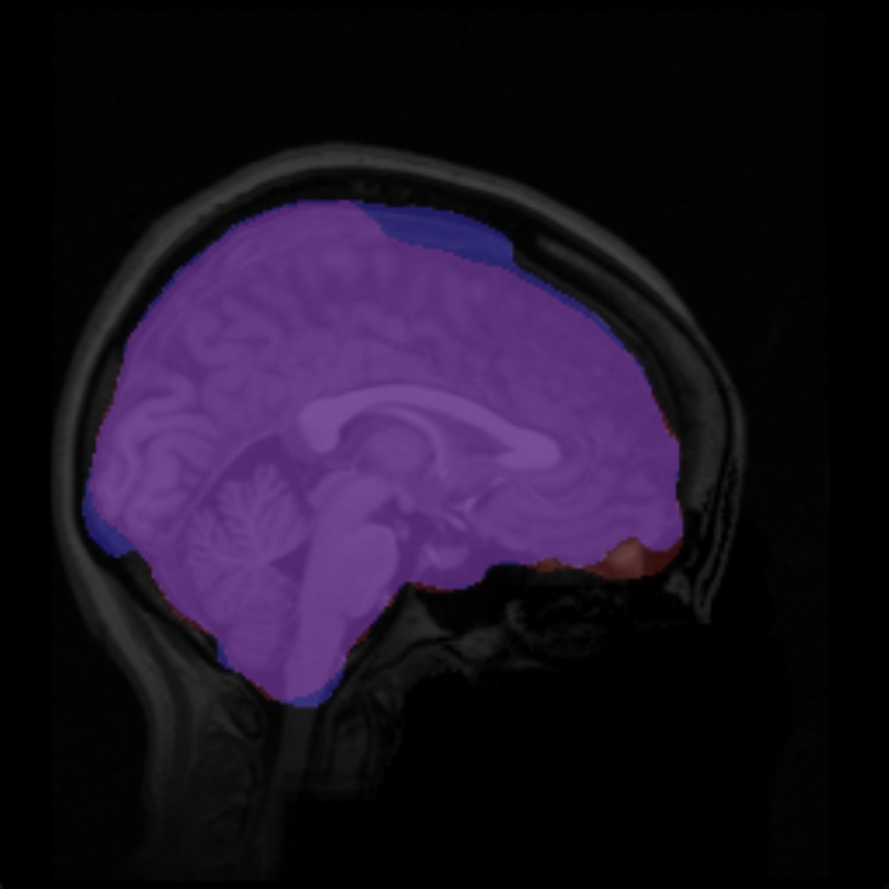}} &
\raisebox{-.5\height}{\includegraphics[width=0.142\columnwidth]{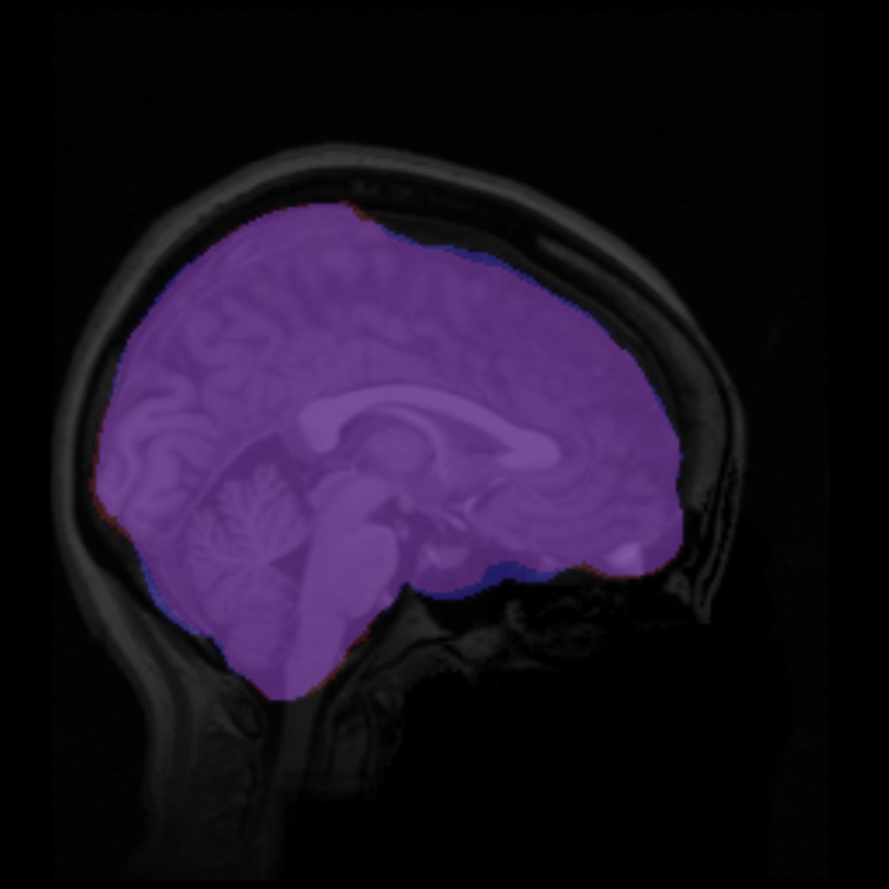}} &
\raisebox{-.5\height}{\includegraphics[width=0.142\columnwidth]{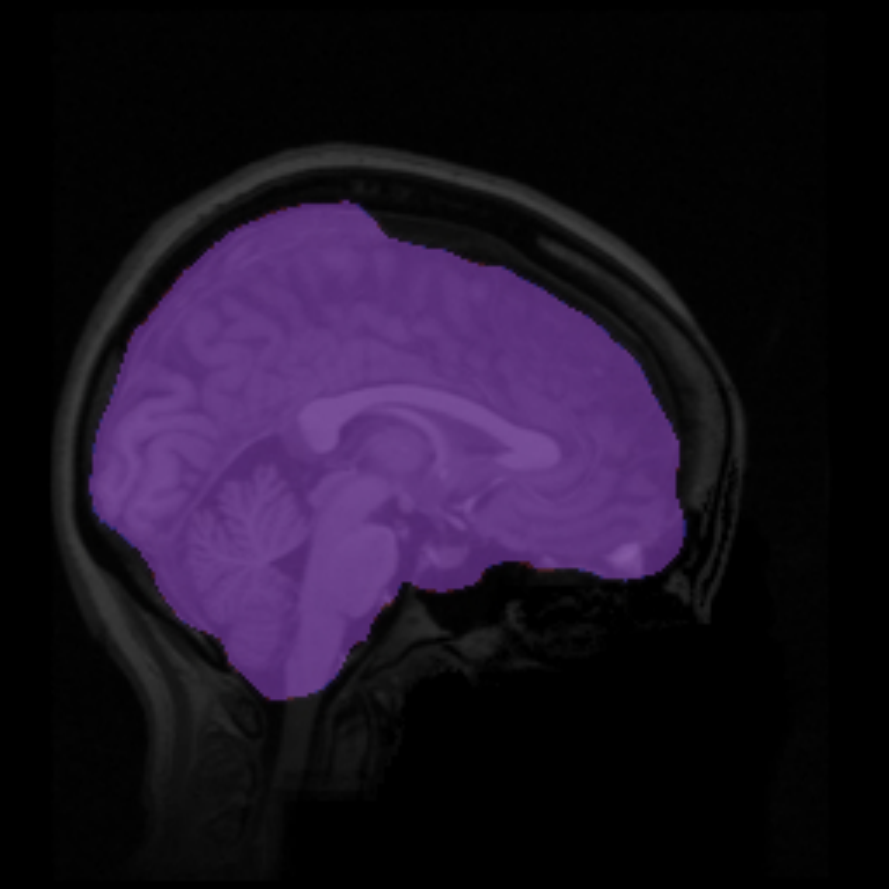}} \\
\small{Case 1} & 
\raisebox{-.5\height}{\includegraphics[width=0.142\columnwidth]{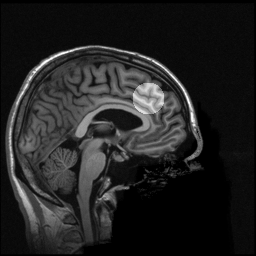}} &
\raisebox{-.5\height}{\includegraphics[width=0.142\columnwidth]{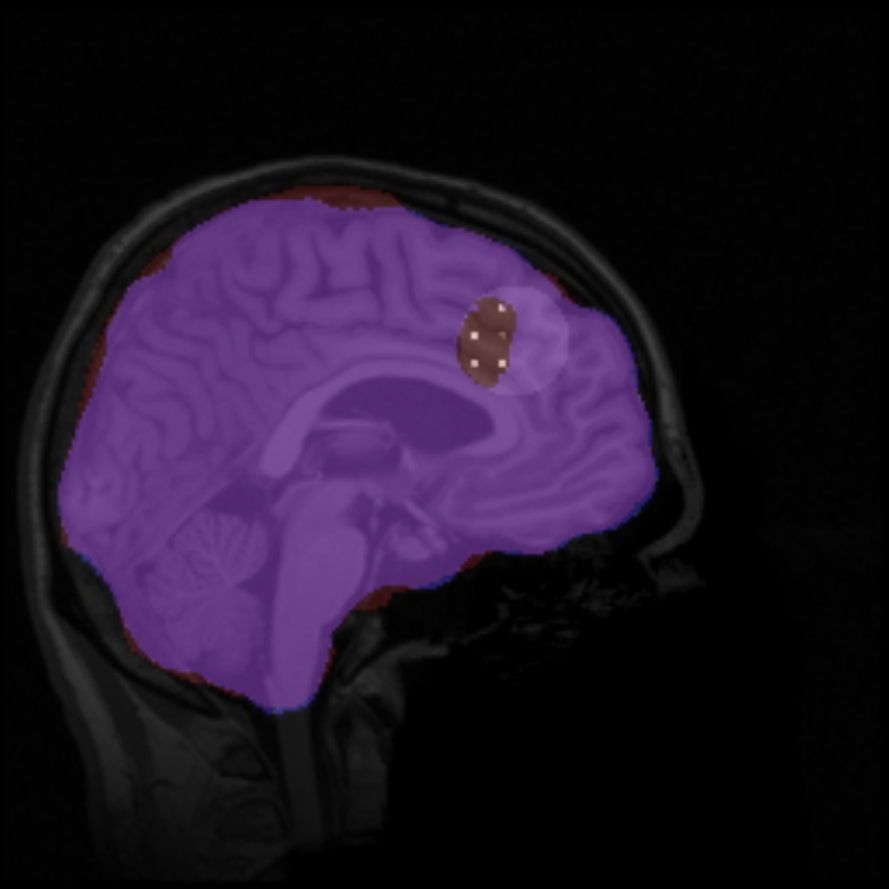}} &
\raisebox{-.5\height}{\includegraphics[width=0.142\columnwidth]{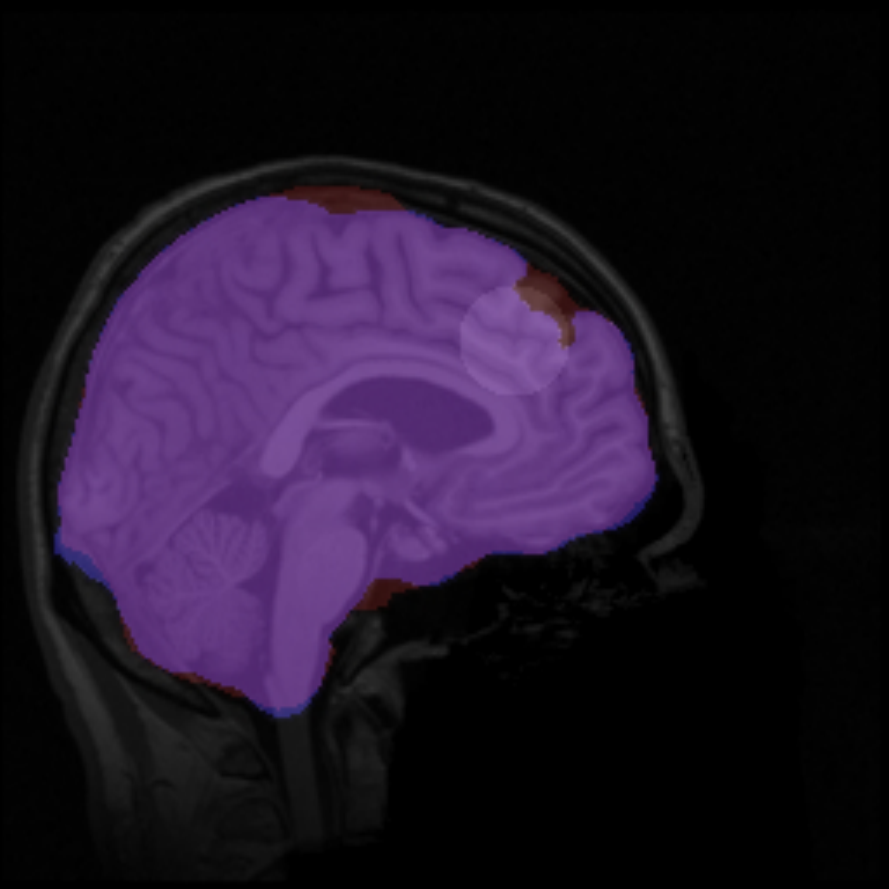}} &
\raisebox{-.5\height}{\includegraphics[width=0.142\columnwidth]{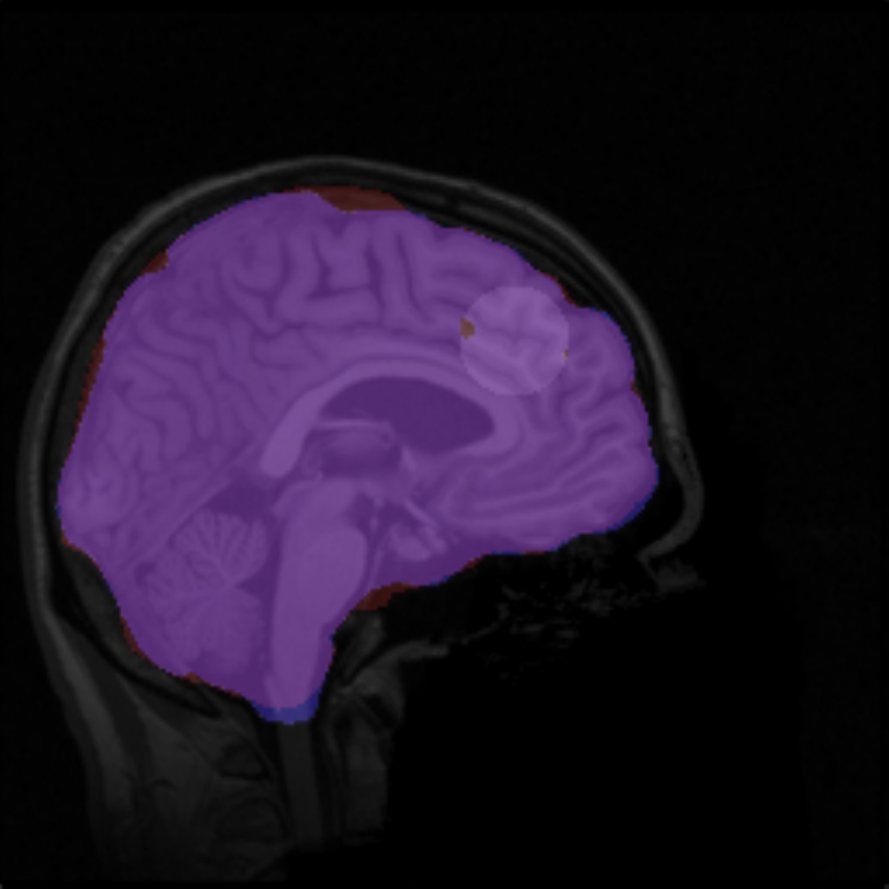}} &
\raisebox{-.5\height}{\includegraphics[width=0.142\columnwidth]{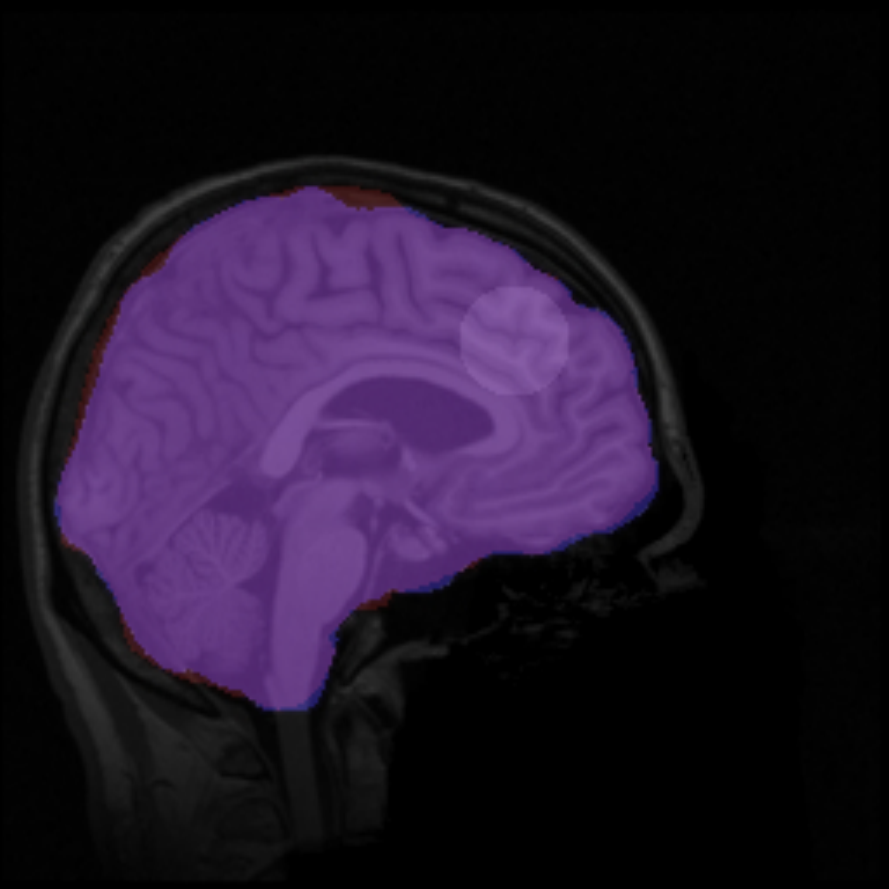}} &
\raisebox{-.5\height}{\includegraphics[width=0.142\columnwidth]{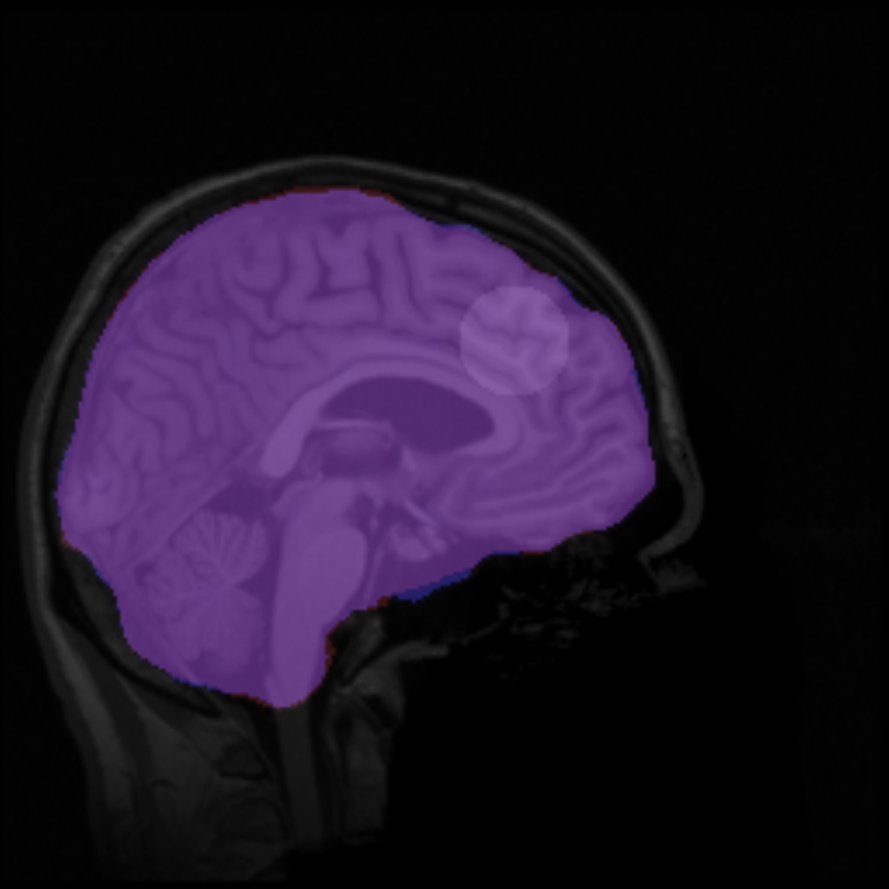}} \\
\small{Case 2} & 
\raisebox{-.5\height}{\includegraphics[width=0.142\columnwidth]{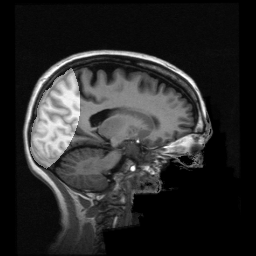}} &
\raisebox{-.5\height}{\includegraphics[width=0.142\columnwidth]{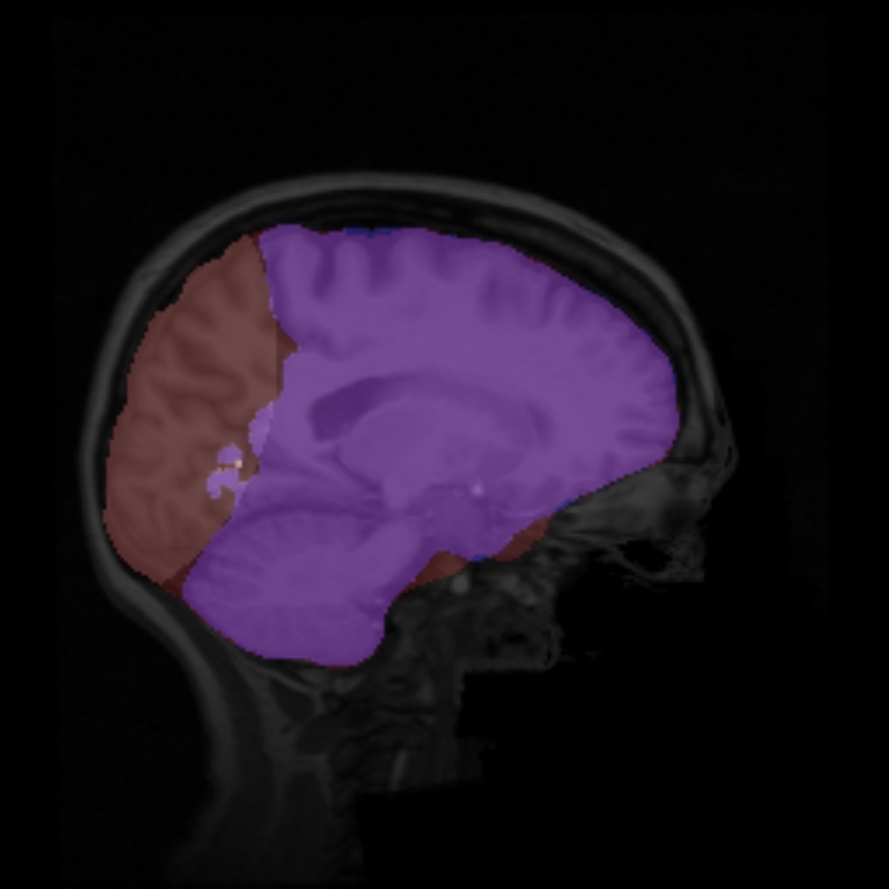}} &
\raisebox{-.5\height}{\includegraphics[width=0.142\columnwidth]{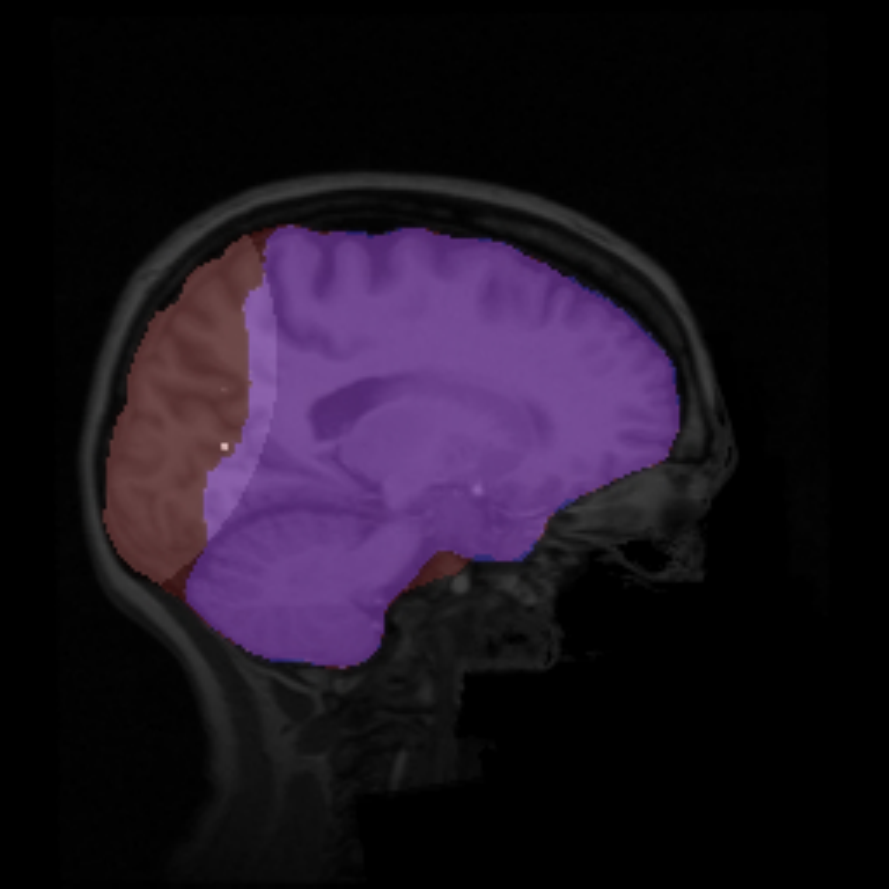}} &
\raisebox{-.5\height}{\includegraphics[width=0.142\columnwidth]{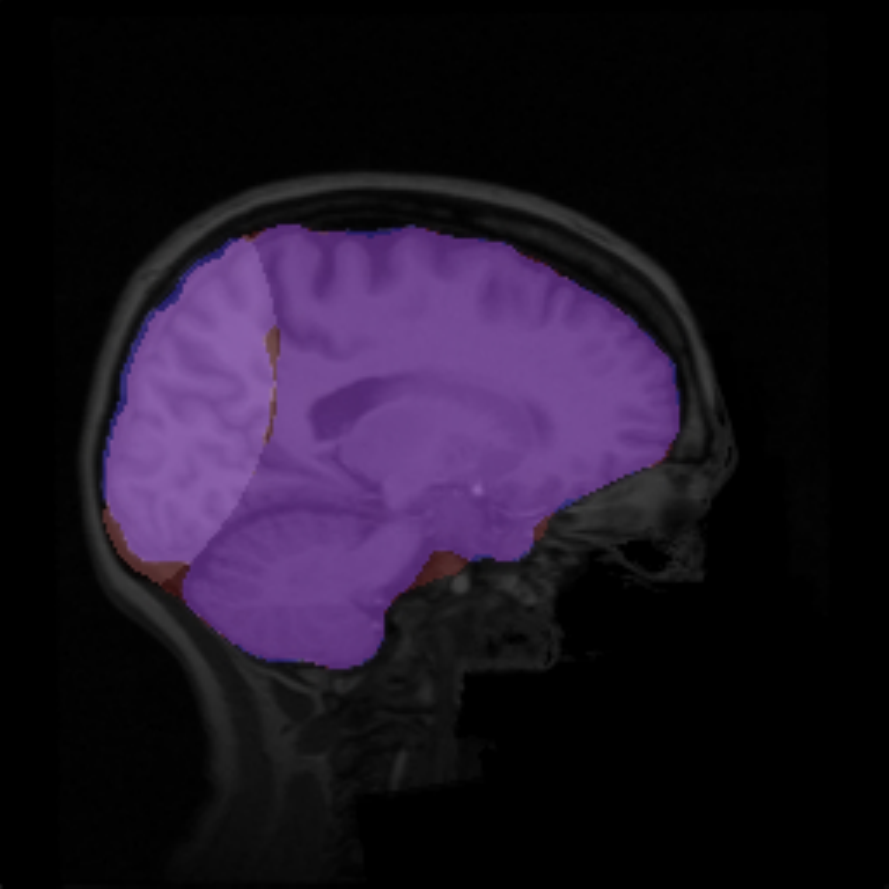}} &
\raisebox{-.5\height}{\includegraphics[width=0.142\columnwidth]{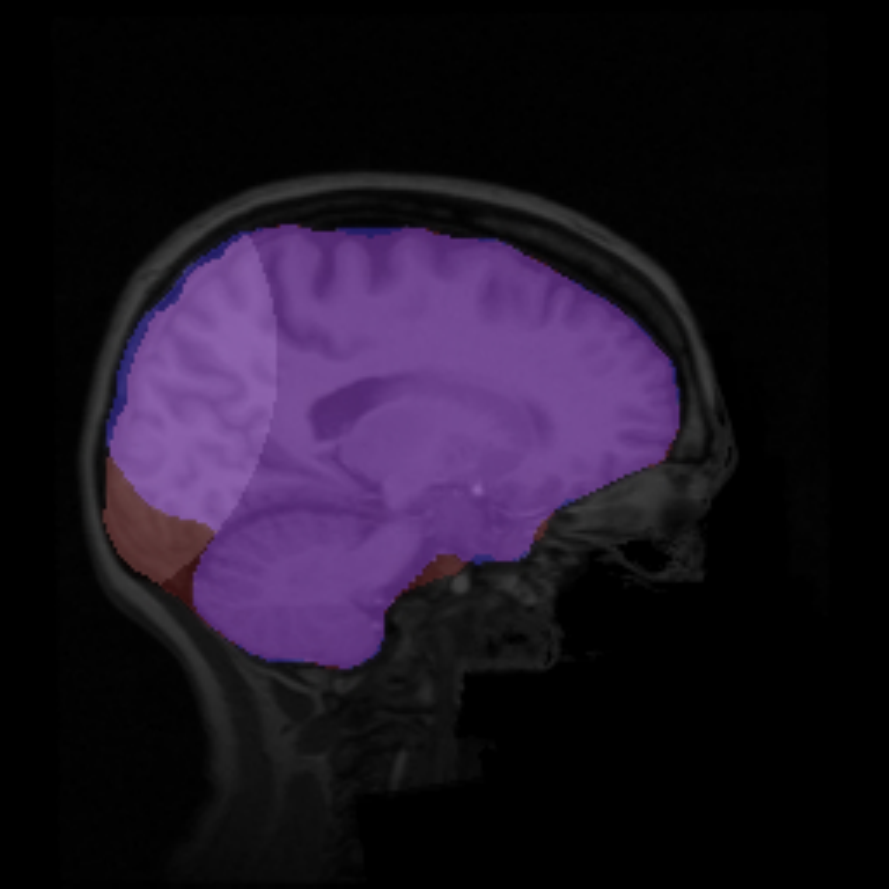}} &
\raisebox{-.5\height}{\includegraphics[width=0.142\columnwidth]{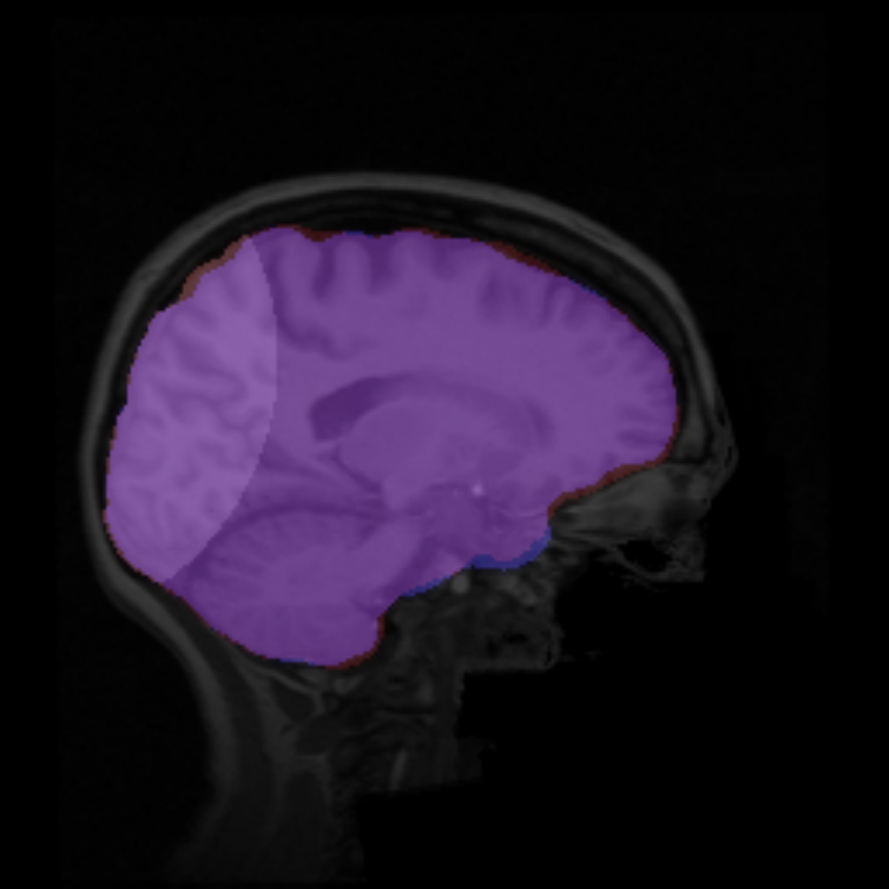}} \\
\small{Case 3} &
\raisebox{-.5\height}{\includegraphics[width=0.142\columnwidth]{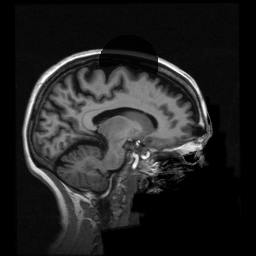}} &
\raisebox{-.5\height}{\includegraphics[width=0.142\columnwidth]{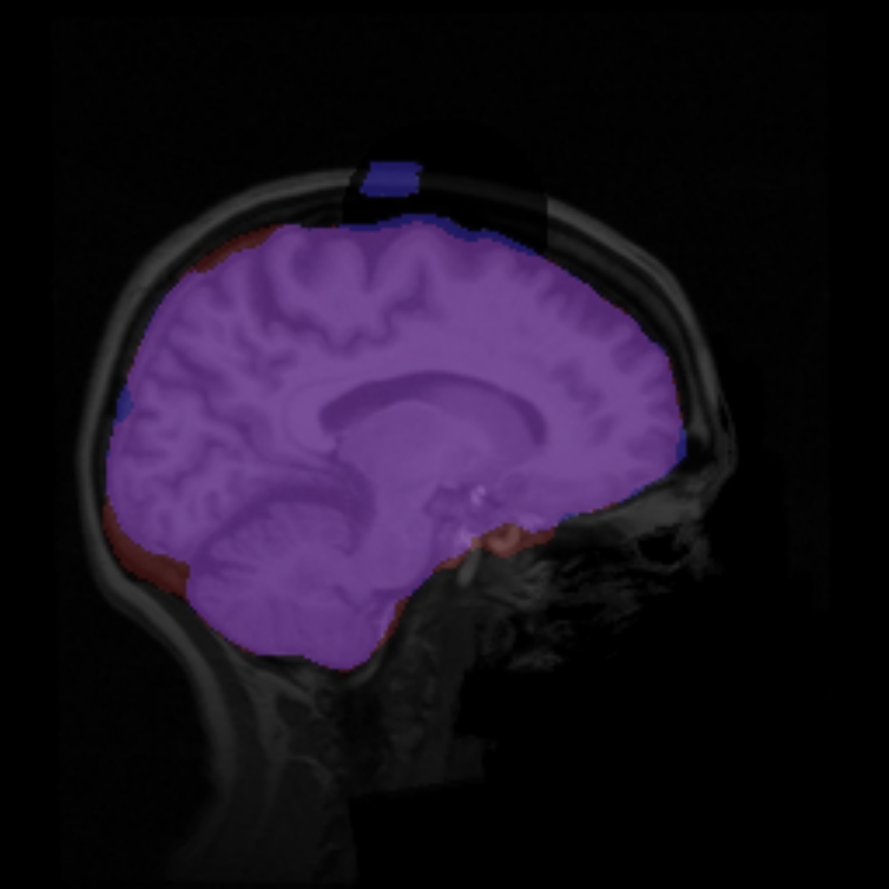}} &
\raisebox{-.5\height}{\includegraphics[width=0.142\columnwidth]{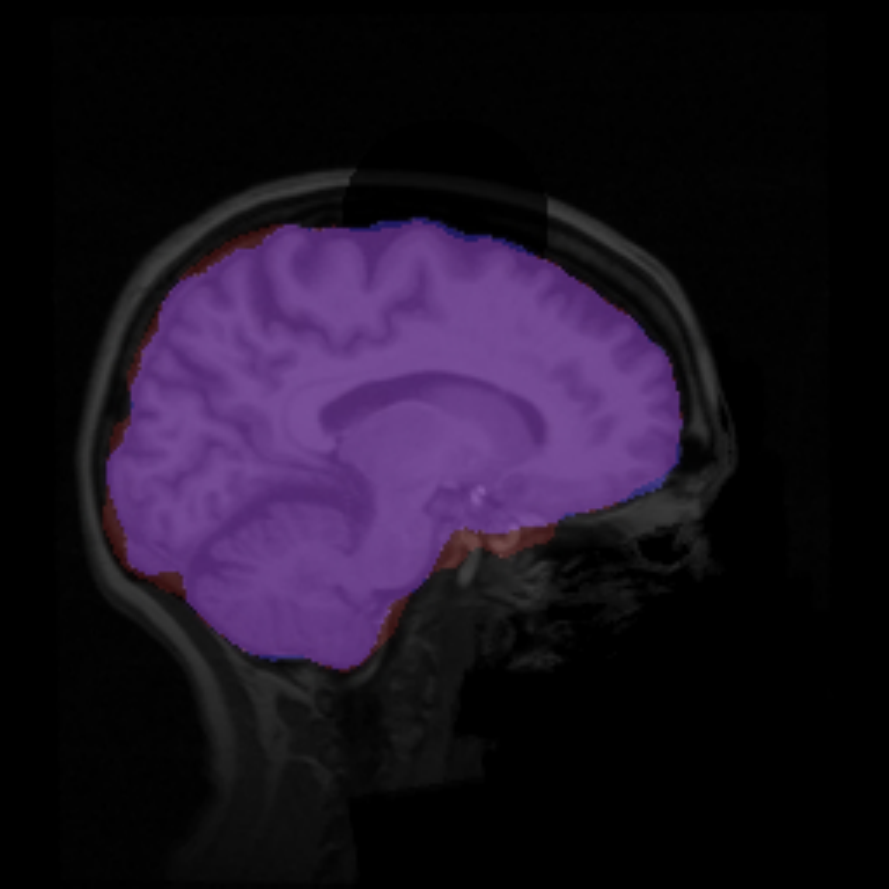}} &
\raisebox{-.5\height}{\includegraphics[width=0.142\columnwidth]{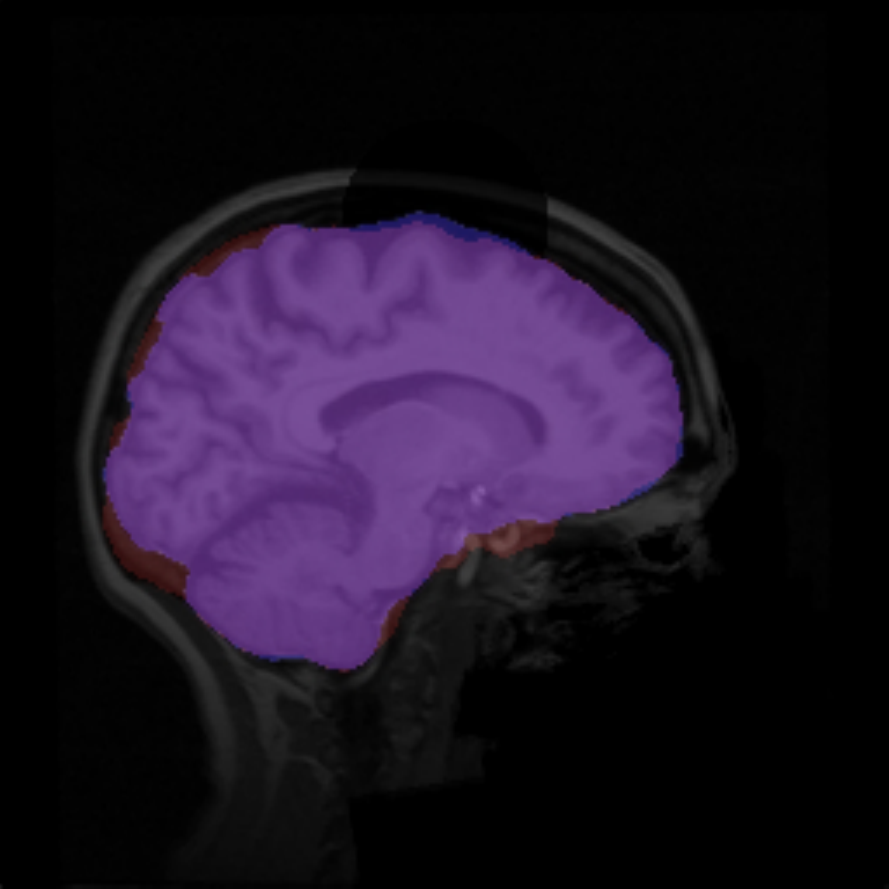}} &
\raisebox{-.5\height}{\includegraphics[width=0.142\columnwidth]{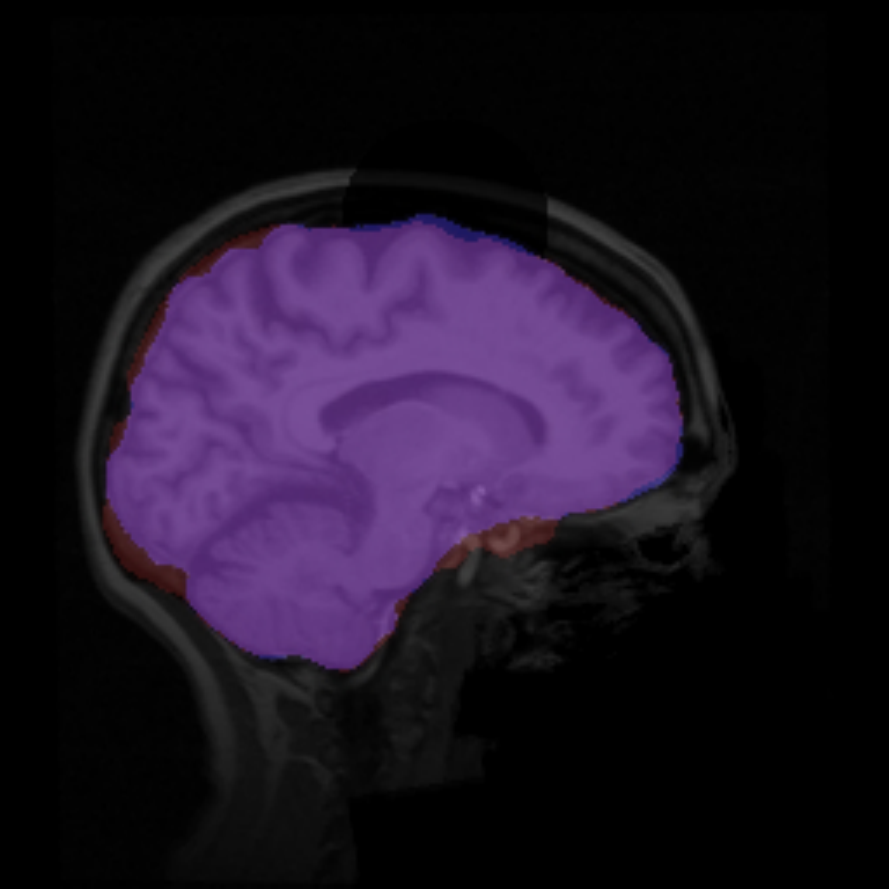}} &
\raisebox{-.5\height}{\includegraphics[width=0.142\columnwidth]{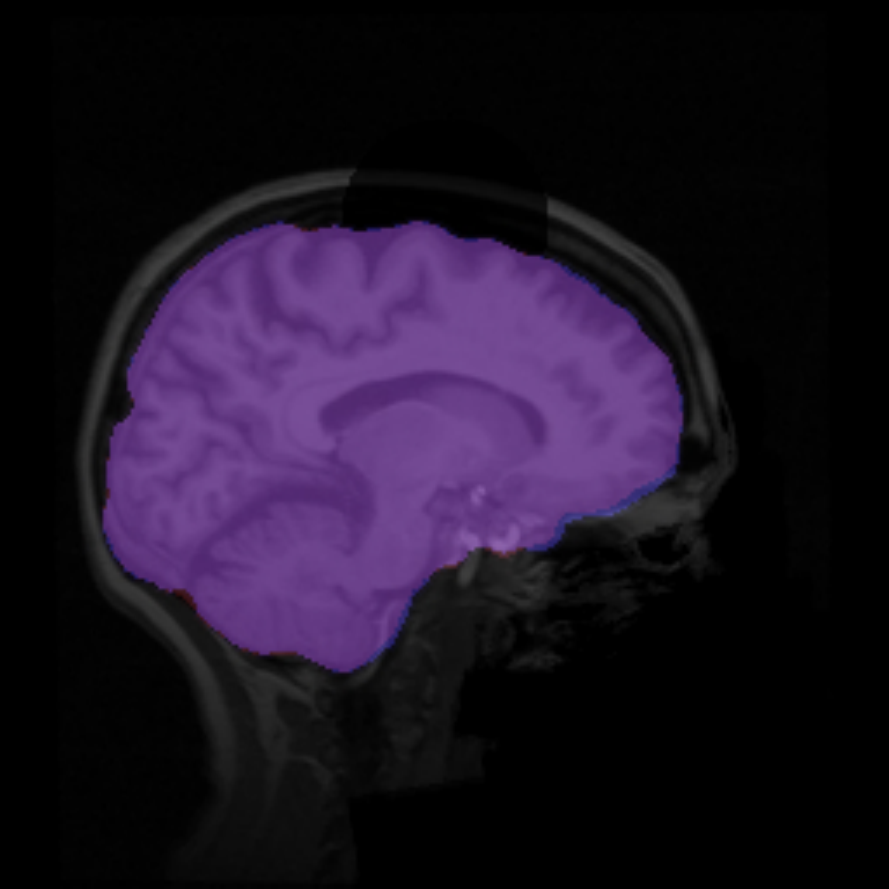}} 
\end{tabular}
\caption{Qualitative comparison among five networks, plain and dense U-Nets, probability, plain, and optimal CompNets, on four image samples: a normal one, one with pathology inside of the brain (case 1), one with pathology on the boundary of the brain (case 2), and one with a damaged skull (case 3). The true (red) and predicted (blue) masks are superimposed over the original images. The purple color indicates a perfect overlap between the ground truth and the prediction. Best viewed in color.} 
\label{fig:visual_results}
\end{figure}

To further evaluate the robustness of our networks, in one chunk of the OASIS subset we introduce brain pathologies, such as synthetic 3D brain tumors and lesions with different intensity distributions, into the images at different locations and with different sizes, as well as damaged skulls and non-brain tissue membranes, as shown in the first column of Fig.~\ref{fig:visual_results}. We train networks on the other chunk of unchanged images and test them on this chunk of noisy images.


\begin{table}[t]
\centering
\begin{tabular}{l|lll|lll}
\footnotesize
& \multicolumn{3}{c|}{(Apparently) Normal Images} & \multicolumn{3}{c}{Pathological Images} \\
& \multicolumn{1}{c}{Dice} &  \multicolumn{1}{c}{Sensitivity} & \multicolumn{1}{c|}{Specificity} & \multicolumn{1}{c}{Dice} &  \multicolumn{1}{c}{Sensitivity}  & \multicolumn{1}{c}{Specificity}  \\
\hline
{Kleesiek\tiny{~et al.}\scriptsize{\cite{kleesiek2016deep}$^\star$}} & {95.77}\scriptsize{$\pm$0.01} & {94.25}\scriptsize{$\pm$0.03} & {99.36}\scriptsize{$\pm$0.003} & \multicolumn{1}{c}{--} & \multicolumn{1}{c}{--} & \multicolumn{1}{c}{--} \\
{Plain U-Net} & {92.30}\scriptsize{$\pm$6.20} & {95.60}\scriptsize{$\pm$1.48} & {96.20}\scriptsize{$\pm$0.09} & {79.90}\scriptsize{$\pm$8.10} & {93.80}\scriptsize{$\pm$5.10} & {95.20}\scriptsize{$\pm$2.15} \\
{Dense U-Net} & {96.40}\scriptsize{$\pm$4.10} & {97.50}\scriptsize{$\pm$0.70} & {96.90}\scriptsize{$\pm$0.01} & {85.43}\scriptsize{$\pm$5.80} & {96.13}\scriptsize{$\pm$3.20} & {97.10}\scriptsize{$\pm$1.27} \\
{Prob. CompNet} & {95.10}\scriptsize{$\pm$0.19} & {96.73}\scriptsize{$\pm$0.90} & {96.03}\scriptsize{$\pm$0.02} & {92.10}\scriptsize{$\pm$5.23} & {96.32}\scriptsize{$\pm$1.90} & {98.86}\scriptsize{$\pm$0.50} \\
{Plain CompNet} & {96.70}\scriptsize{$\pm$0.22} & {97.93}\scriptsize{$\pm$0.62} & {98.57}\scriptsize{$\pm$0.06} & {95.21}\scriptsize{$\pm$3.75} & {96.32}\scriptsize{$\pm$1.03} & {99.21}\scriptsize{$\pm$0.10} \\
{Opti. CompNet} & \cellcolor{green!10}{{98.27}\scriptsize{$\pm$0.30}}~ & \cellcolor{green!10}{{98.26}\scriptsize{$\pm$0.58}}~ & \cellcolor{green!10}{{99.80}\scriptsize{$\pm$0.05}} & \cellcolor{green!10}{{97.62}\scriptsize{$\pm$2.21}}~ & \cellcolor{green!10}{{97.84}\scriptsize{$\pm$0.80}}~ & \cellcolor{green!10}{{99.76}\scriptsize{$\pm$0.12}}
\end{tabular}
\caption{Quantitative comparison (mean and standard deviation in percentage) among different networks on (apparently) normal and pathological images. $^\star$This paper is not directly comparable to our networks, because it was evaluated on mixed data samples, including 77 images ($57\%$) from OASIS data set. (Prob.: Probability; Opti.: Optimal) }
\label{tab:quantitative_results}
\end{table}

\paragraph{\bf Experimental Settings.}
Apart from dropout with a rate of 0.3, we also use $L_2$ regularizer to penalize network weights with large magnitudes and its control hyperparameter $\lambda$ is set to 2e-4. For training, we use the Adam optimizer~\cite{kingma2014adam} with a learning rate 1e-3. All networks run up to 10 epochs.     

\paragraph{\bf Experimental Results.} We compare our CompNets with a 3D deep network proposed in~\cite{kleesiek2016deep}, a plain U-Net (the backbone of the plain CompNet), and a dense U-Net (the backbone of the optimal CompNet). These networks are tested on (apparently) normal images (with two-fold cross-validation) and on pathological images (with being trained on the other fold with clean images). Given a 3D brain MRI scan of a subject, our networks accept 2D slices and predict brain masks slice by slice, which are stacked back to a 3D mask without any post-processing. 

Figure~\ref{fig:visual_results} demonstrates the qualitative comparison among predicted brain masks. For (apparently) normal brain scans, all networks produce visually acceptable brain masks. However, the plain and dense U-Nets have difficulties in handling images with pathologies, especially the pathological tissue on or near the boundary of the brain. Part of the pathological tissue in the brain is considered as non-brain tissue. The plain U-Net even oversegments part of the skull as the brain when the skull intensity changes, as shown in the case 3. In contrast, our CompNets can correctly recognize the brain, and the optimal CompNet presents the best visual results for all four cases. We then use Dice score, sensitivity, and specificity to quantify the segmentation performance of each network, as reported in Table~\ref{tab:quantitative_results}. The optimal CompNet consistently performs the best among all networks for either normal (averaged Dice of $98.27\%$) or pathological (averaged Dice $97.62\%$) images, although its performance on images with pathologies is slightly downgraded by $<0.7\%$ on average and $<2.6\%$ in the worst case.

\begin{figure}[t]
\centering
\begin{tabular}{cccc}
\quad \, \includegraphics[width=0.18\columnwidth]{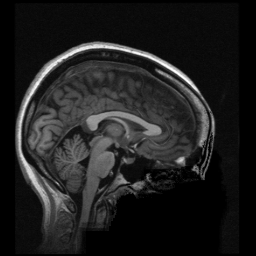} \, \quad& 
\quad \, \includegraphics[width=0.18\columnwidth]{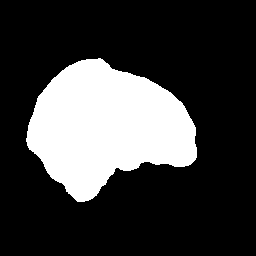} \, \quad &
\quad \, \includegraphics[width=0.18\columnwidth]{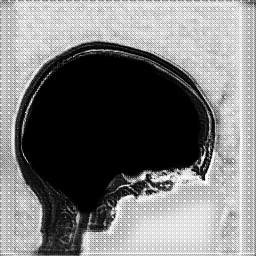} \, \quad &
\quad \, \includegraphics[width=0.18\columnwidth]{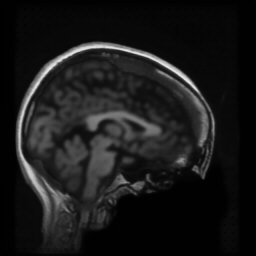} \, \quad \\
\small{(a) Input} & \small{(b) Brain mask} & \small{(c) Complement} & \small{(d) Reconstruction}
\end{tabular}
\caption{Three outputs (b-d) of our optimal CompNet for an input brain scan (a).} 
\label{fig:optimal_compnet_outputs}
\end{figure}

Figure~\ref{fig:optimal_compnet_outputs} shows the three outputs from the optimal CompNet: the masks for the brain and its complement and the reconstructed image. According to the combination of the brain mask and its complementary one, we can identify different parts in the brain image. This confirms that the brain branch works as expected; more importantly, the complementary branch has learned features for separating the non-brain region from the brain tissue. This enables the network to handle unseen brain tissue and be insensitive to pathologies in brain scans.  


\section{Discussion and Conclusions}
\label{sec:discussionAndConclusions} 
In this paper, we proposed a complementary network architecture to segment the brain from an MRI scan. We observed that the complimentary segmentation branch of the optimal CompNet learns a mask outside of the brain and can help recognize different structures in the non-brain region. The complimentary design makes our network insensitive to pathologies in brain scans and helps segment the brain correctly. We used synthetic pathological images due to the lack of publicly available brain scans with both pathologies and skulls. Our source code is publicly available for in-house testing on real pathological images~\footnote{\url{https://github.com/raun1/Complementary_Segmentation_Network.git}}. 

Furthermore, our current networks accept 2D slices from a 3D brain image, but the design can be extended to 3D networks for directly handling 3D images. Implementing 3D CompNets will be one of our future work plans. In addition, our complementary network design is not specific to the brain extraction problem but can be generalized to other image segmentation problems if the complementary part helps learn and understand the objects of interest. Another future work is the analysis of the theoretical and geometric implications of our CompNets. 


\bibliographystyle{splncs03}
\bibliography{brain-extraction}

\begin{thebibliography}{10}
\providecommand{\url}[1]{\texttt{#1}}
\providecommand{\urlprefix}{URL }

\bibitem{dey2018diagnostic}
Dey, R., Lu, Z., Hong, Y.: Diagnostic classification of lung nodules using 3{D}
  neural networks. In: International Symposium on Biomedical Imaging. pp.
  774--778 (2018)

\bibitem{dice1945measures}
Dice, L.R.: Measures of the amount of ecologic association between species.
  Ecology  26(3),  297--302 (1945)

\bibitem{han2017brain}
Han, X., Kwitt, R., Aylward, S., Menze, B., Asturias, A., Vespa, P., Van~Horn,
  J., Niethammer, M.: Brain extraction from normal and pathological images: A
  joint {PCA}/image-reconstruction approach. arXiv:1711.05702  (2017)

\bibitem{huang2017densely}
Huang, G., Liu, Z., Weinberger, K.Q., van~der Maaten, L.: Densely connected
  convolutional networks. In: CVPR. vol.~1, p.~3 (2017)

\bibitem{ioffe2015batch}
Ioffe, S., Szegedy, C.: Batch normalization: Accelerating deep network training
  by reducing internal covariate shift. In: ICML. pp. 448--456 (2015)

\bibitem{kingma2014adam}
Kingma, D.P., Ba, J.: Adam: A method for stochastic optimization.
  arXiv:1412.6980  (2014)

\bibitem{kleesiek2016deep}
Kleesiek, J., Urban, G., Hubert, A., Schwarz, D., Maier-Hein, K., Bendszus, M.,
  Biller, A.: Deep {MRI} brain extraction: a 3{D} convolutional neural network
  for skull stripping. NeuroImage  129,  460--469 (2016)

\bibitem{marcus2007open}
Marcus, D.S., Wang, T.H., Parker, J., Csernansky, J.G., Morris, J.C., Buckner,
  R.L.: Open access series of imaging studies ({OASIS}): cross-sectional {MRI}
  data in young, middle aged, nondemented, and demented older adults. Journal
  of cognitive neuroscience  19(9),  1498--1507 (2007)

\bibitem{ronneberger2015u}
Ronneberger, O., Fischer, P., Brox, T.: U-net: Convolutional networks for
  biomedical image segmentation. In: International Conference on Medical image
  computing and computer-assisted intervention. pp. 234--241. Springer (2015)

\bibitem{salehi2017auto}
Salehi, S.S.M., Erdogmus, D., Gholipour, A.: Auto-context convolutional neural
  network for geometry-independent brain extraction in magnetic resonance
  imaging. arXiv:1703.02083  (2017)

\bibitem{souza2017open}
Souza, R., Lucena, O., Garrafa, J., Gobbi, D., Saluzzi, M., Appenzeller, S.,
  et~al.: An open, multi-vendor, multi-field-strength brain {MR} dataset and
  analysis of publicly available skull stripping methods agreement. NeuroImage
  (2017)

\bibitem{srivastava2014dropout}
Srivastava, N., Hinton, G., Krizhevsky, A., Sutskever, I., Salakhutdinov, R.:
  Dropout: A simple way to prevent neural networks from overfitting. The
  Journal of Machine Learning Research  15(1),  1929--1958 (2014)

\bibitem{xia2017w}
Xia, X., Kulis, B.: W-net: A deep model for fully unsupervised image
  segmentation. arXiv:1711.08506  (2017)

\end{thebibliography}

\end{document}